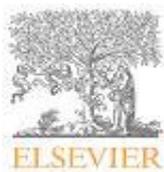

# Software Based Higher Order Structural Foot Abnormality Detection Using Image Processing

Arnesh Sen, Kaustav Sen, Jayoti Das[*]

*Department of Physics, Jadavpur University, Kolkata 700032, India*

**Abstract**

The entire movement of human body undergoes through a periodic process named Gait Cycle. The structure of human foot is the key element to complete the cycle successfully. Abnormality of this foot structure is an alarming form of congenital disorder which results a classification based on the geometry of the human foot print image. Image processing is one of the most efficient way to determine a number of footprint parameter to detect the severeness of disorder. This paper aims to detect the Flatfoot and High Arch foot abnormalities using one of the footprint parameters named Modified Brucken Index by biomedical image processing.

## 1. Introduction

Brucken Index[1][2] of the human footprint is the ratio of the series of the lines connected between the medical border lines to the lateral border lines. The other existing footprint parameters are arch index, arch angle, footprint index, arch length index, truncated arch index[1][2]. By analyzing of an imprint of foot these indexes are introduced to calculate the ratio of the foot contact area to the remaining area in different ways. Brucken Index of the human footprint is the ratio of the series of the lines connected between the medical border lines to the lateral border lines. By analyzing of an image of the foot these indexes are introduced to calculate the ratio of the foot contact area to the remaining area in different ways.

The main objective is to determine in what percentage of foot arch covered in the entire area. If this percentage was high with respect to some upper threshold value then flat foot was considered and if this percentage was lower than some lower threshold value then high arch abnormality was considered. The percentage in between the upper and lower threshold was considered as normal arch. This logic was maintained and the percentage is calculated based upon the above mentioned indexes to differentiate between Flat foot [3], Normal Foot and High Arch Foot. In the question of reliability of this existing technique, problems are occurred to detect higher order abnormality. This logic can be utilized only to determine the type of abnormalities but it failed to determine how serious the abnormality is. In the existing technique the imprint of the foot was cropped as close as the shape of the footprint. For case of normal arch a noncontact area is surely comes

———
[*] Corresponding author: Jayoti Das, Tel.8967976134; email: jayoti.das@gmail.com

out and for the case of flat foot this noncontact area is decreased but not vanished.

If the noncontact area is vanished for the extreme flat arch on can unable to find the footprint parameters as those parameters are nothing just the ratio of the contact area to the total area. Alternatively, for the case of extreme high arch case one cannot extract the imprint accurately as the major portion of the arch is not in contact with the floor.

In this paper the existing techniques can be improved by some modification with introduction of a new foot print index named "Modified Brucken Index" which defines the ratio of the foot-floor contact area, measured in terms of pixels, to the non-contact area also measured in terms of pixels. It should be noted that in MBI, pixels are used instead of series of lines. Therefore the errors will certainly be minimized in the proposed process. The process of cropping the image of footprint will also be reexamined to detect the higher order flat foot abnormalities. In the case of determination of higher order high arch abnormalities the new logic suggests that the snap shot of the side view of the foot arch should be taken and examined by image processing technique to solve the problems occurred in previous or existing logic.

## 2. Methodology

According to the proposed logic, to find modified Brucken Index medical border line is set to the edge of a rectangular shaped plain glass of constant ratio and this area should sufficient enough to hold human foot. Here 21 cm x 27.9 cm rectangular glass plain is used. Now the snap shot is taken from the opposite side and the image will be cropped to extract the curvature part of the foot. Two strong light sources are used to eliminate the shadow part and untouched curved portion according to the figure-4. The cropped ratio is 1/3$^{rd}$ [4] of the foot axis as assumed to calculate the Arch Index in conventional method. According to the feasibility test it is also noticed that the ratio of the total pixel area of the foot to the area contains the curvature part is almost 0.34. Now the cropped image is converted into its binary form with the help of image processing [5]. The binary form will return the image in the form of an array whose dimension is depend on the pixel resolution of the image. The foot contact region will be considered as the dark portion of the image and returns binary "0" and the remaining portion will be considered as bright part or binary "1". The illuminated untouched curved portion is treated as bright part of the image and will be considered as binary '1". Thus we have an array of a particular dimension consist of only 0 and 1. If we assumed the series of lines using in Brucken Index have no gap between them, then we of course talk about not only about the lines we talk about the whole area. Now in the image processing the binary '0's and binary '1's represents dark and white pixels respectively. The resolution of the foot print image should be too high to allow us to assume the total number of pixels are almost equal to the area of the image. Thus total number of '0's and '1's represent the area, occupied by the footprint and the remaining area of the glass plain respectively. The ratio of these two quantity may be termed as the Modified Brucken Index.

### 3.1 Higher Order Flat-Foot Abnormality Detection

To detect higher order flat foot abnormality snap shot of the foot is taken with that glass plain that cropped image will be processed with image processing method as previous way. Hence the modified Brucken Index of the image can be calculated. In this way the modified Brucken Index will vary according to the higher order flat-foot abnormality [6].

### 3.2 Higher Order High-Arch Abnormality Detection

For extreme high arch abnormality detection the first step is to check if the patient is suffering from extreme high arch problem[7][8] or not and the second step is to detect the order of the abnormality. In the first step if the image of the foot is continuous, the Modified Brucken Index of the foot is evaluated in a same way, discussed in previous case. This abnormality is considered as first order abnormality.

## 3. Experimental Setup

According to Figure-3 this experimental set up consist of one hollow platform like inverted backless chair, one transparent fiber or glass sheet [9] which area is larger than A4 size, one high resolution camera [10](DSLR is the best choice) and two strong light source. This fiber or glass sheet, on a hollow platform acts as a transparent floor. The backless chair is oriented in inverted position and the fiber or glass sheet is kept on it. Now two strong light sources are fixed in either side of that transparent platform as a way that they can able to illuminate the foot. The camera is fixed vertically beneath that platform as it can take the photograph of the foot to platform contact area. The entire experimental setup is explained clearly by Figure-1. After taking the snap shot the image will be processed.

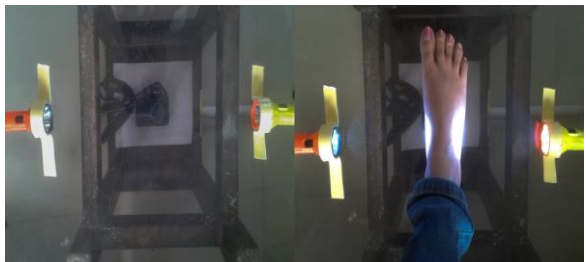

Figure-1

## 4. The Image Processing Technique

With the help of suitable software now the image of the foot will be processed. For this modelling [11] MATLAB is used for this task. For this experimental setup the background colors may cause a severe error at time of converting the image into the binary form. The software cannot able to differentiate between the various color profiles. It will convert the actual image to the binary image depending upon the pixel intensity of the various color. Thus it will also convert the background into the binary image as in Figure-2. As a result the image of the foot to floor contact area,[12] as well as the background is now converted to binary and MBI will be evaluated includes the background. Thus for evaluating the exact value of MBI we have to get rid of the interference of the background.

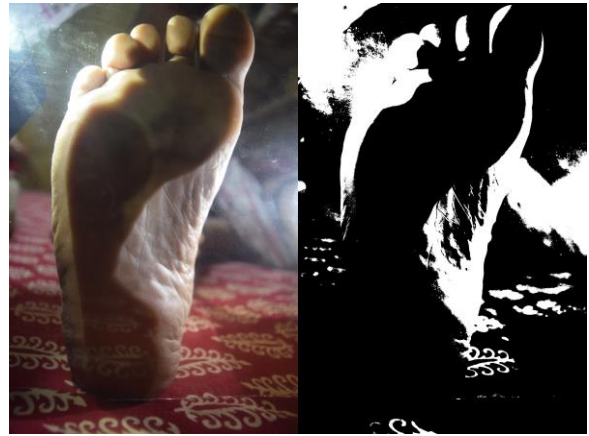

Figure-2

Now to eliminate the background a white cover can be used as white is converted to binary '1' and the contact area which is dark with respect to the background is converted to binary '0' as expected. But till now this technique [13] is not totally error free. It is impossible to cover entire the background as a space is needed to allow the human foot to put on the floor. As a result the error caused by the background will not be totally eliminated but minimized. To get more accuracy a new intermediate step is introduced where the image is converted to the grey scale [14]-[16] with respect to a particular threshold value and then it is converted to binary image. This threshold value is chosen as a way that it is always greater than the pixel intensity of the contact area and less than any other undesired illuminated pixel intensity.

## 5. Choice Of Threshold Value

Here the threshold value is chosen at 140 which means the pixel's intensity above 140 is treated as while or binary '1' and below 140 is treated as black or binary '0'. At this threshold value the contact area is almost undisturbed. In the figures below, Figure-7 explains how the image is distorted from the desired one. In this experiment this threshold value is standardize and all the feasibility test was done with respect to this threshold. Now to check the reliability of this threshold a graph is plotted in Figure against the various threshold value and the percentage error of MBI. In Figure-3 first one explains that if Threshold Level is decreased significantly from the tolerance value then how the image is distorted and 2$^{nd}$ one explains the opposite situation which means how image will be distorted if the threshold value is significantly increased from the tolerance level.

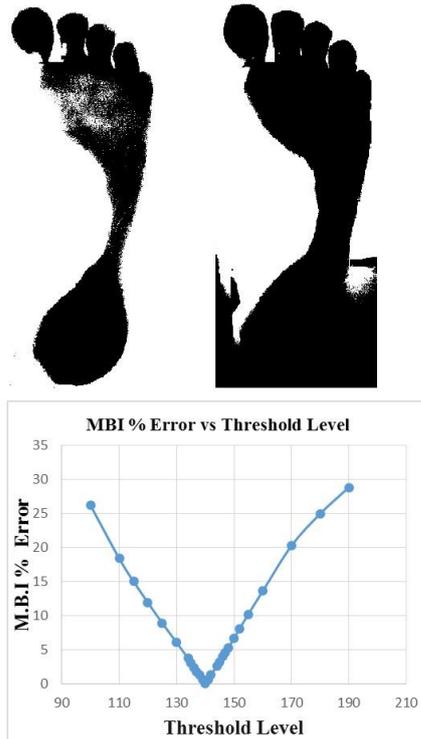

Figure-3

From the above in Figure 3 we can explain how the slight change of threshold value influences the percentage error of M.B.I . According to this plot for 7% tolerance of threshold value maximum percentage error occurs is 5% and for 15% tolerance the maximum error occurs is 10%.

## 6. Experimental Results

In this experiment all the foot images are taken through the above mentioned method and suitable distance from this A4 sheet glass/fiber and camera is 30 cm. All the snapshots are taken vertically and if the experimental results give that the existence of the high arch abnormality then snap shots are taken also from the side view with a distance of 30 cm also. Here digital single lens reflex (DSLR) camera is used with FINE mode as this mode takes the image in JPG format with highest resolution. As our logic demands the highest possible number of pixels to minimize the error the FINE mode is the best choice. Then the image is cropped and processed. From Table-1 we can conclude that if different types of foot images (Normal Foot, Flat Foot, and High Arch Foot respectively) are examined M.B.I will vary according to the abnormalities. Now this experiment is repeated again and again on different foot images to find how the exact range of M.B.I will differ between these abnormalities and how the order of the abnormality will be detected by M.B.I.

Table-1

Variation of M.B.I according to Foot Structure.

| RESOLUTIONS (DPI) | PIXELS | NUMBER OF BINARY "1" (B1) | NUMBER OF BINARY "0" (B0) | M.B.I (B1/B0) | FOOT TYPE |
|---|---|---|---|---|---|
| **300** | 3841 x 4442 | 4094736 | 3304980 | 1.23896 | Normal Foot |
| **300** | 3841 x 4442 | 2414504 | 6734858 | 0.35851 | Flat Foot |
| **300** | 3841 x 4442 | 6154872 | 2994490 | 2.05540 | High Ankle Foot |

## 7. Feasibility Test

To find the range of the curved structure of human foot an ample number of foot structure is investigated to find the ratio between the arch length covered by the curvature part and total arch length which is almost 0.34 [18]. That's why one can calculate the curvature part which is the main area of interest of this research by measuring the arch length. To determine the MBI, this experiment is applied to the 70 people suffering from flat foot [19]-[21] abnormalities, 150 people of normal foot structure and 40 people suffering from high arch abnormality. Experimental results shows that people have normal foot structure the value of MBI varies within 0.7 to 1.4. People, suffering from flat foot abnormality the value of MBI lies between 0.1 to 0.38 and those who are affected by high arch abnormality MBI lies between 1.7 to 2.3. By this process the order of the high arch abnormality detection is not possible. For this we have to follow the 2$^{nd}$ process as discussed in Methodology section.

If the modified bracken index is plotted (Figure-4) against the area of the non-contact portion of the A4 sheet which is approximated by the pixels, one can get three region. Patients, those MBI is lying at the middle region of the graph suffering from flat foot abnormalities. The higher region and lower region of the graph indicates the MBI of those, who are suffering from flat foot [22] and high arch foot abnormalities.

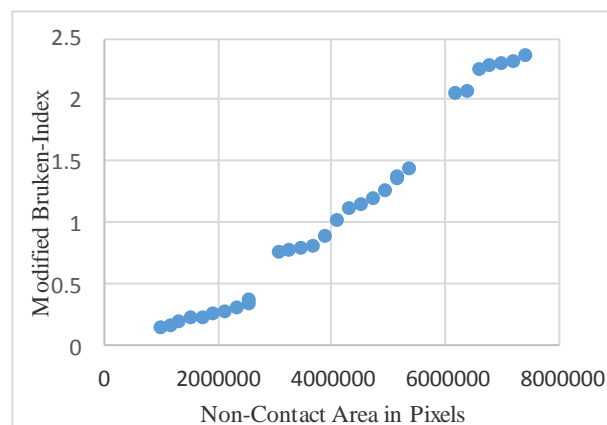

Figure -4

## 8. Conclusion

The process is no doubt more advance and it minimizes the error as compare to the conventional technique of finding BI. In this experiment the total number of pixels is assumed as the area of the image. That's why by increasing the resolution of the image as much as possible the error can be minimized. Moreover the conventional method is depend on the imprint of the foot. The method of taking imprint for determining the conventional Brucken Index involves a number of errors depending on the ink swallowing and paper quality. The present method is able to minimize these kind of errors and the process is far more efficient and may be considered as a medical aid to solve the higher order structural foot abnormality very easily.